\documentclass{article}
\usepackage{gensymb}
\usepackage{authblk}
\usepackage{lscape} 
\usepackage{tabularx}
\usepackage{moreverb,url}
\usepackage[english]{babel}
\usepackage{amsfonts}
\usepackage{mathtools} 
\usepackage{amssymb}
\usepackage{amsmath}
\usepackage{geometry}
\usepackage{cite}
\usepackage{graphicx}

\title{A review paper of bio-inspired environmental adaptive and precisely maneuverable soft robots}
\author{Mengqi Shen \ mqshen@vt.edu}
\begin{document}

\maketitle

\begin{abstract}
This paper summarizes the most recent research in soft robotic field from material, actuation, mechanics property, dimension $\&$ scale and architecture. The presents application
relations among the functionalities, manufacturing process and the factors mentioned above.
\end{abstract}


\section{Introduction}

Traditionally, robots made by rigid materials have been widely used in manufacturing. However, the lack of flexibility and absorption of energy causes the interaction between machines and humans to be quite dangerous. In contrast, the deformability, adaptivity, sensitivity and agility enable soft robots to better bridge this gap. Unlike the rigid robots whose movement can be described as 6 degrees of discrete freedom (3 rotations and 3 translations about the x, y, z axes), the intrinsic deformation of soft robots is continuous, complex and compliant, which is considered offering infinite degrees of freedoms\cite{R1}. Thus, it’s extremely hard to control the motion of robots made by soft materials directly through reverse and inverse kinematics. As a result, quantifying and replicating the behavior of soft materials becomes one of the major challenges. With the development of modern techniques, such as microscale 3D printing, this challenge can be solved by: 
\begin{enumerate}
\item Functional materials. Designing novel materials with locally tuned composites, tailored microstructures and desired properties. The responses will be directly programmed into the materials\cite{R4} itself. One example of such materials is metamaterial.
\item Forcing the designs to simple tasks, such as bending\cite{R2} gripping\cite{R41} and crawling\cite{R3}. By assembling the basic types of single motion(bending, twisting or translation) to achieve reliable and predictable performance\cite{R5}.
\item Designing geometrical shapes or architectures whose functional response is programmed within the structures. Such as stimuli-responsive lattice structure\cite{R4}\cite{R7}, origami and kirigami structure\cite{R13}, textiles\cite{R33,R34,R35,R36,R37,R38}
\item Dynamic simulation of soft multi-materials soft objects\cite{R15}.
\end{enumerate}

The flexibility and functionality of soft materials and architectures enable the infinite possibility of soft robots. Hence, the goal of the soft robotic is to conduct multiscale research on novel soft materials and soft robots-- Under the micro scale, investigate novel metamaterials with locally tailored composites, intricated microstructure and tuned properties; Under meso- to macro- scale quantifies the overall kinematics of the soft materials and structures, integrate different functionalities of the different materials, then to design complex soft robotic architecture on demand to the required environment (such as autonomous soft robot with sensor). And the application of such soft robots will be autonomous shape-morphing robots/wearable medical- assistance robot/bio-mimicking autonomous robot that are capable of navigating, sensing, tracking and executing demands under unknown environments.
\section{Methodology}
Soft materials are usually capable of large deformation and mechanical response to the external stimulus, and such stimulus could be force, temperature, electric field and pressurized fluid, etc. The diversity of the type of responses (e.g. contraction, expansion, vibration, elasticity, etc) of soft materials provides us with infinite probability to create the soft robots on demands to different kinds of tasks. The key to creating such robots lies in the design of the materials, actuators, mechanical responses, scale, dimension and architecture. Hence, the first step to realize the ultimate goal of this proposal is to investigate and summarize those key factors directly from cutting-edged research.
\subsection{Material}
Soft robotic systems are fundamentally soft and highly deformable[20]. In order for the body of the soft robots to achieve its potential for sensing, rescuing, disaster response, and human assistance, it requires compliance of its material. \\

Elastomers are the most common flow-modulus material, which includes: Silicone rubber: low modulus that allows high strain and the convenience of room temperature vulcanizing process\cite{R8};Hydrogel: highly stretchable and tough\cite{R9}, can be manufactured in various shape without requiring support materials\cite{R10}; Polyurethanes: strong hydrogen bonding between the polymer chains, which enables polyurethanes to exist as solid polymers at room temperature, toxicity and flammability concerns remain\cite{R10}; Dissolvable materials: can be used in the medical field. \\

The continuum robots are not necessarily made by soft materials. For example, many worm-like robots are used shape-memory alloys (SMA) for their large force-weight ratio, large life cycle, and noise-free operation\cite{R11}\cite{R21}. \\

Similar to SMA, soft memory polymer (SMP) can maintain two or more stable configurations under different conditions\cite{R4,R13,R43}. Generally, there are three types of SMP and most of them are either thermo- or chemo- responsive\cite{R27}. \\

Liquid metals and highly ductile metals compose of flexible conductive material family. For example, Ge-In alloy is a eutectic metal with a broad temperature range of the liquid phase\cite{R42}. Reducing the thickness of copper or gold and embedding them into the deformable structure can achieve soft electronics\cite{R2,R40}. \\

Inspired by the biological materials made by living organisms in nature, there are increasingly attentions on the novel materials (e.g. metamaterials) with locally tailored composition, intricate microstructures, and tuned properties which cannot be fabricated by conventional routes. For example, elastic-modulus-controllable printable ink[4]; Locally programmed bubble size, volume fraction and connectivity polymer foam\cite{R22}; Auxetic beam\cite{R39}. Liquid Crystal with spatially programmed nematic order\cite{R23}.
\subsection{Actuation}
Each material has its unique strain/stress, deformation, and velocity response to the stimulus from the environment. The transduction methods of temperature, pressurized fluid, electrical field and tension cables are the most commonly used. According to the different transduction methods, we can classify the actuation as follows. \\

Thermal actuation: The material system can expand or contract in response to the temperature. Interweaving materials with different thermal expansion and elastic modulus can achieve controlling the underlying metric tensor in space and time\cite{R4}. The kinematic is predictable under different temperature, thus it’s capable of creating a soft robot that can repeatedly shape-morph and propel\cite{R13}. \\

Pneumatic/hydraulic (fluid) actuation: Fluid-driven actuators are simple, with large actuation stress and strain, low cost and fast response times. A linear contraction and large force can be produced when positive fluidic pressure is inflated into the channels, thus it requires high-pressure fluid to achieve desired force and displacement. The air is the most common pressurized fluid used in the pneumatic actuator. The pneumatic actuators require valves to precisely control their locomotion\cite{R30}, which substantially consumes energy. Electropermanent magnet (EPM) valve can easily switch the state of flux output simply driven by small pulsed current ($5ms$)\cite{R31}. Vacuums have also been utilized to create strong contraction with minimal changes in actuator radius\cite{R32} or change the states between flowing and rigid of the granular fillers\cite{R41}. The phase-changing materials that are used as pressurized fluid can sustain the deformation\cite{R26}. \\

Electrically actuation: Since energy is typically stored in electrical form, it may be more efficient to directly use electrical potential to actuate soft robots. This type of actuation is based on the polymeric matrices activated with a different mechanism. Types of electroactive polymers (EAP) include dielectric EAPs (DEA), ferroelectric polymers, liquid crystal polymers, ionic EAPs, electrorheological fluids, ionic polymer-metal composites, and stimuli-responsive gels\cite{R1}. Electroactive polymer composites such as ionic polymer-metal composites (IPMCs) are activated under small voltage. An example of DEA is ionic hydrogel-elastomer hybrid, which is formed by laminating layers of hydrogels and elastomers together. It requires a large electric field to deform the elastomeric material. The dielectric elastomers layer will exhibit in-plane expansion and compression along its thickness[12]. Some configurations can couple both planar directions into a single out-of-plane direction\cite{R29}. \\

Tendinous actuation: Such actuation is achieved by embedding soft segments with the anisotropic mechanical behavior of tendons (e.g. tension cable, inextensible fiber\cite{R32}, cellulose yarn\cite{R33}, SMA). A typical mechanism is the incorporation of the strain-limiter, such as the precise control of the deformation trajectory of the soft actuator\cite{R32}. Artificial muscles made of conducting polymer actuators were assembled by textile processing cellulose yarns coated with conducting polymer[33]. Varying the angles of knitted fibers and the selective contracting of individual fibers grant the ability to move fiber muscle precisely and with an infinite degree of freedom\cite{R34}. 

\subsection{Mechanics property}
Pneumatic actuators are typically fabricated by elastomeric materials, such as the EcoFlex family,
with the elastic strains greater than $600\%$, The result of tensile tests indicate that the average elastic
modulus is $5.5Mpa$, the average ultimate strength is about $2.5Mpa$. The average toughness is about $5kJm^{-3}$,, and elastomers typically failed after 6 to 12 cycles[24]. The pressurization leads to the large-range twisting and contractile motions. Maximal torsion is 225$\degree$ and the maximal contractor is  $8.5\%$ at the pressure of $6kPa$ and actuation speed of up to $18mm/s$\cite{R7}. \\

Shape memory alloys (SMA) can memorize their previous form when subjected to stimuli such as thermomechanical or magnetic variations\cite{R27}. Hysteresis and nonlinearity make SMAs difficult to be precisely controlled. The Young’s modulus of SMA families vary from $28GPa$ to $41GPa$, and yield strength varies around $70MPa$ to $140MPa$. The SMA can generate large contractile stress that is over $200MPa$ as an artificial muscle\cite{R28}. \\

SMPs are relatively easy to manufacture and fast to train. Comparing to SMA, SMP has a lower cost, better efficiency and biodegradable. Its mechanical properties by far may surpass SMA. In addition, SMPs can sustain two or more changes (e.g. bistability) when triggered by thermal\cite{R4,R13,R27}. The critical elastic modulus (before phase transition temperature) for SMPs varies from $0.01GPa$ to $3Gpa$. The deform stress varies from $1MPa$ to $3MPa$. Recovery stress also varies from $1MPa$ to $3MPa$. Low-cost polymer fibers such as twisted fishing line and sewing thread, can generate large stress up to $140MPa$ and tensile strokes up to $49\%$\cite{R28}. \\

IPMC actuator is repeatable, suits for low-frequency driving input. The deformation of the IPMC actuator is typically small since it is driven by the expansion of the cathode, thus it is applicable in the scenario that requires moderate deformation. IPMCs actuator can consume large current though it can be driven with small voltage ($<5V$) \cite{R36,R36}. DEAs are driven by large electric field ($> 1KV$) with fast response time and low power requirement. EAPs are resilient materials with high failure strains ($>300\%$). The elastic modulus of EAPs is smaller than $1MPa$ ($0.1MPa$ to $1MPa$\cite{R12}). \\

The actuation mechanism in conducting polymer (CPs) is dominated by mass transfer, which includes ions and solvents into the polymer. CPs can generate large stress, by parallel assembling the fibers coated by CP, the total force of the actuator can increase while the high surface-to-volume ratio will be maintained. The strain of the individual yarn will increase vary from $0.075\%$ to $0.3\%$ by changing the different yarn materials\cite{R32}. 

\subsection{Dimension and Scale}
Efficient, power-dense and energy-efficient actuators are what we are looking for to power the soft robots. The functionality of the soft robotics systems relies on their material and structural properties. From the single degree of freedom which assembles the basic actuation of soft robot, to the robot that has mobility and sensing in the 2D plane by combing each locomotion together\cite{R25}. The phase-changing materials with microfluidic origami structure allow the 2D laminates to evolve into a third spatial dimension\cite{R26}. And finally, the robot can self-propel\cite{R13}, shape-morphing\cite{R4}, directional control under a fourth dimension\cite{R43}. \\

The metamaterials have drawn increasing attention due to their unique mechanical properties that are controlled by their locally tunable geometries (unit cells). The manipulation of the microscale cell promises the desired properties (e.g. density\cite{R22}, frequency\cite{R4}, modulus\cite{R4}, Poisson’s ratio\cite{R37}, etc) of the materials under the macroscale. Though there has been a tremendous number of research on macroscale soft robots, such as artificial muscle(\cite{R31,R32,R33,R34,R35}), soft robotic fish[44]. The soft functional devices under the micro- and meso- scale remain exploring, due to where the fabrication of complex structures still remain challenging. For example, a soft robotic spider in which the minimum cutting distance is about $40 \mu m$  that is driven by a microfluidic actuator\cite{R26}. The smart composite microstructure enables a centimeter inchworm that can achieve difficult locomotion such as crawling and climbing\cite{R45}.
\subsection{Architecture}
Origami and kirigami-based structures: Origami and kirigami are the arts of folding and cutting to create things with as little processing as possible. The combination of origami and kirigami with smart material actuators enables the designing of intricate actuation with less complexity\cite{R37}. The origami and kirigami approach to make robots can be considered as a top-down approach. That is fabricating in-plane structures and then folding them along the predefined creases to form 3D objects. Thus, origami and kirigami-based strategies offer the design of soft robots that exhibit large shape changes through a reduced set of predictable motions. The conventional origami designs are not able to carry mechanical loads, however, an origami-inspired architecture of the polymeric tiles interconnected by SMP (liquid crystal elastomer) active hinge can adopt 3 different stable configurations depending on its exposure to the different thermal conditions[13]. By merging well-established techniques such as multilayer soft lithography, bulk micromachining and origami folding, a soft microstructure with increased functional and structural complexity was created. The exploitation of phase-changing materials allows the 2D laminates to evolve into a third spatial dimension\cite{R26}. An example of kirigami soft robots is the kirigami skin with a single pneumatic actuator to achieve better crawling ability by mimicking the extensibility, stretchability and anisotropic frictional properties of the snake skin\cite{R38}.  \\

Beam-based structures: The beam-based elements are widely used for programming deformation patterns since they can be easily manufactured by 3D printing. A variety of mechanical properties can be achieved by the carefully arrangement of the beam elements. For example, by manipulating the unit beam under the microscale, the auxetic behavior (negative Poisson’s ratio) can be observed\cite{R39}. The elastic beam made of SMP can snap between two or three configurations, which can be exploited to realize the large locomotion trigged by temperature, as demonstrated by two programmed, untethered, self-propulsion without a battery or onboard electronics robots\cite{R1,R43}. The elastomeric beam can generate high bending motion\cite{R12}, which makes it the perfect candidate to mimic human muscle. In the meanwhile, the elastomeric beam also presents some reversible nonlinear behaviors such as buckling, distortion and collapse of a set of interconnected beams under negative pressure, such reconfiguration can be used to produce a range of motions with single input[46] that are similar to mammalian muscle\cite{R47}. \\

Reinforced architectures: The soft materials share an infinite degree of freedom and isotropic mechanical behavior (such as elastic modulus) which make it unprecedently convoluted to impart the movements. However, combing with the stimuli-responsive reinforcements of anisotropic mechanical behavior, the motion can be programmed along specific directions. The Stretchable Adhesive Uni-directional prepreg (STAUD-prepreg) can effectively create unique strain-limiting patterns and govern the shape when adhering to volumetrically expanding soft bodies\cite{R32}. The artificial skin controlled by paracord can apply surface strain and pressure to sculpt the clay into the desired shape\cite{R48}. The reinforcements’ orientations of self-shaping composites are controlled by the weak external magnetic field, combing such orientation control with the deformation ability of polymer matrices, it is possible to program specific shape changes under microstructure scale\cite{R49}. \\

Voxel structures: Traditional simulation methods such as finite element analysis no longer fit in simulating the large deformation of soft materials, due to the stiffness of the mesh and nonlinearity of the system equations. However, limiting the discretized elements to voxel can solve such a problem. The force and stiffness of each constituent element can be efficiently computed\cite{R50}. And a self-damage-recovery soft robot based on the voxel model has also been created\cite{R15}. \\

Ultrathin structures: The wearable and sensing devices are important in soft robotics due to their mechanically flexible and robust. Reducing the thickness of highly ductile and conductive metals such as gold enables their uses in constructing wearable devices, such as high sensitive sensor\cite{R51}. The approach of fabricating ultrathin, deformable and conductive material and then transferring them to the target surface for a conformal interface, has been successful in creating bio-integrated electrical circuits\cite{R52}. By replacing the rigid coil with liquid metal we can obtain the soft electromagnetic actuator with mechanically stable and powerful driven properties\cite{R42}.   \\

\section{Applications}
The research about arbitrary combinations of the functional materials, scales, complexities, structures and responses for now merely cover the potentiality of the soft robots. From the simplest robots that can execute single motion, such as biomimetic earthworm and cockroach, to the robots that capable of shape morphing and propelling, moving and sensing, to the autonomous robots that can escape, self-recover from the injury. The functionality of soft robots is becoming more and more intricate. And it is possible to integrate the all of these characteristics together to build the fully autonomous soft robot that can sensing and adaptive controlling under robust environment. \\

The table below summarizes some typical examples of the soft robot besides their actuations, structures, materials and fabrications. 
\begin{landscape}
\begin{table}[h]
\small\sf\centering
\caption{Summarization of typical soft robots example.\label{T1}}
\begin{tabularx}{\linewidth}{|l|X|l|X|l|X|l|X|}
\hline
Actuation Type & Example & Material & Structure & Mechanism & Fabrication\\
\hline
Pneumatic/Hydraulic actuation & Peacock Spider\cite{R26} &	UV-curable resin &	Origami &	Microfluid and origami &	Lithography, micromachining \\ & Damage recovery robot\cite{R15} &	Silicone rubber &	Voxel &	Pneumatic actuation & 2-axis molding\cite{R17} \\
& Buckling Beam\cite{R47} &	Elastomer &	Beam &	VAMP & \ \\
\hline
Thermal actuation &Shape-morphing lattice\cite{R4} &	PDMS &	Beam lattice &	Lattice with different thermal-expansion coefficients &	4D printing (ABG 10000, Aerotech) \\
& Self-propel rollbot[13] &	LCE	& Origami &	Triple-stability &	HOT-DIW \\
& Untethered swimming bot\cite{R43} &	VeroWhitePlus & plastic &	Beam  &	Bistability	\\
\hline
Electrically actuation & Jellyfish bot\cite{R42} &	Ga-In and PDMS &	Ultrathin &	EMA	& Cured process \\
& DEA Actuator\cite{R12} &	Ionic hydrogel-elastomer & Beam & DEA & DIW \\
& Cockroach robot\cite{R2} & Kapton	& Beam &	EAP	& \ \\
\hline
Tendinous actuation & Adhesive composite lamina\cite{32} &	STAUD-prepreg &	Reinforce	&
Tendinous &	Cured and textile processing \\
& Artificial muscle\cite{R33} &	Cellulose yarn and CP &	Reinforce &	
Tendinous &	Textile processing \\
& Self-shaping composite\cite{R49} & CMF &	Beam	&
Magnetic and tendinous	& \
\end{tabularx}
\end{table}
\end{landscape}

\end{document}